\documentclass[a4paper,twoside]{article}

\usepackage{epsfig}
\usepackage{subcaption}
\usepackage{calc}
\usepackage{amssymb}
\usepackage{amstext}
\usepackage{amsmath}
\usepackage{amsthm}
\usepackage{multicol}
\usepackage{pslatex}
\usepackage{apalike}
\usepackage{algorithm2e}
\usepackage[bottom]{footmisc}

\usepackage{multirow}
\usepackage{lineno}
\usepackage{subcaption}
\usepackage{graphicx}
\usepackage{url}

\usepackage{color}

\usepackage{SCITEPRESS}     

\begin{document}

\title{Sleep-stage efficient classification using  a lightweight self-supervised model}

\author{\authorname{Eldiane Borges dos Santos Durães\sup{1} and João Batista Florindo\sup{1}\orcidAuthor{0000-0002-0071-0227}}
\affiliation{\sup{1}Institute of Mathematics, Statistics and Scientific Computing, University of Campinas, Street Sergio Buarque de Holanda, 651, Campinas, Brazil}
\email{e250962@dac.unicamp.br, florindo@unicamp.br}
}


\abstract{\textbf{Background and Objective:} Accurate classification of sleep stages is crucial for diagnosing sleep disorders and automating this process can significantly enhance clinical assessments. This study aims to explore the use of a self-supervised model (more specifically, an adapted version of mulEEG) combined with a Linear SVM classifier to improve sleep stage classification. \textbf{Methods:} The mulEEG model, which learns electroencephalogram signal representations in a self-supervised manner, was simplified here by replacing ResNet-50 with 1D-convolutions used as time series encoder by a ResNet-18 backbone. Two other adaptations were conducted: the first one evaluated different configurations of the model and data volume for training, while the second tested the effectiveness of time series features, spectrogram features, and their concatenation as inputs to a Linear SVM classifier. \textbf{Results:} The results showed that reducing the volume of data offered a better cost-benefit ratio compared to simplifying the model. Using the concatenated features with ResNet-18 also outperformed the linear evaluations of the original mulEEG model, achieving higher classification performance. \textbf{Conclusions:} Simplifying the mulEEG model to extract features and pairing it with a robust classifier leads to more efficient and accurate sleep stage classification. This approach holds promise for improving clinical sleep assessments and can be extended to other biological signal classification tasks.}

\onecolumn \maketitle \normalsize \setcounter{footnote}{0} \vfill

\section{\uppercase{INTRODUCTION}}
\label{sec:introduction}

In the human body, sleep is divided into cycles, each consisting of two distinct phases: Rapid Eye Movement (REM) and Non-Rapid Eye Movement (NREM) sleep, which is further divided into stages N1, N2, and N3 \cite{sleep-stages}. Each phase is characterized by variations in muscle tone, brain wave patterns, and eye movements. Furthermore, the human body undergoes each sleep cycle 4 to 6 times per night, with each cycle lasting approximately 90 minutes. However, sleep quality and time spent at each stage can be affected by sleep-related or mental disorders, such as obstructive sleep apnea, depression, schizophrenia, and dementia, as well as traumatic brain injuries, medications, and circadian rhythm disorders. As a result, the identification of sleep stages plays a crucial role in the diagnosis of these conditions, and automating this classification can significantly improve clinical sleep assessments.


The mulEEG model described in \cite{kumar2022mulEEG} is an example of a modern successful approach in the deep learning category. It aims to automate the classification of sleep stages using electroencephalogram (EEG) signals collected during sleep. To achieve this, it performs a pretext task, which involves learning effective representations of these EEG signals from multiple data views in a self-supervised manner. Despite the great results achieved by the mulEEG model in ideal scenarios, with large amounts of data and computational resources for pretraining, we notice that the literature still lacks self-supervised architectures adapted for contexts where the availability of data and computational power is limited.

In this context, and inspired by self-supervised models for time series analysis like mulEEG, here we propose a self-supervised deep learning framework for the identification of sleep stages. We take mullEEG as our starting point, but leverage it with strategies aiming at offering alternatives to simplify both the model and its training process, while also maximizing the utility of the learned representations by using them as input to a Linear SVM algorithm for the classification task.

The first strategy involves replacing ResNet-50 with 1D-convolutions in the time series encoder with a ResNet-18, and alternating the usage of the pretext dataset between $20\%$ and $100\%$. The second one involves several training of the Linear SVM using the representations learned by the self-supervised module with either ResNet-50 or ResNet-18, which are the time series features, spectrogram features, and their concatenation. 

To assess the performance of the proposed model, the main evaluation metrics were accuracy (Acc), Cohen's kappa ($\kappa$), and macro-averaged F1 score (MF1). The reference for comparison was the linear evaluation of mulEEG presented in \cite{kumar2022mulEEG}, where a linear classifier was trained using only the time series features as input. However, the training of the Linear SVM with the concatenation of both features learned by mulEEG using ResNet-18 was sufficient to surpass the results reported in \cite{kumar2022mulEEG}.

\section{RELATED WORKS}

As stated in \cite{SEKKAL2022103751}, the methods used to automate the classification of sleep stages are based on two main strategies: (i) conventional machine learning methods and (ii) deep learning approaches based on artificial neural networks. The first category includes algorithms such as K-Nearest Neighbors (KNN), Support Vector Machines (SVM), Random Forests (RF), Decision Trees, and Bayesian rule-based classifiers. The second category comprises Multi-Layer Perceptrons (MLP) and their modern refinements, including Recurrent Neural Networks (RNNs), Long Short-Term Memory Networks (LSTMs), Gated Recurrent Units (GRUs), Bi-directional LSTMs, and Convolutional Neural Networks (CNNs). Our study lies in the second category. A recent review on deep learning techniques used for sleep stage classification can be found in \cite{liu2024automatic}.

Deep learning can be further divided into different learning paradigms, such as supervised, unsupervised, and, more recently, self-supervised learning. This last group is particularly interesting in scenarios where the access to annotated data for training is limited. Several works have investigated the use of self-supervised learning on EEG signals, for example, the study in \cite{xiao2024self}, where an algorithm based on contrastive learning is used for seizure detection. Specifically on sleep stage classification, we have \cite{eldele2023self}, which provides a systematic evaluation of SSL in few-label settings. The authors in \cite{yuan2024self} opt for analyzing non-polysomnography data, particularly acquired by wrist-worn accelerometers. In our study, we focus specifically on mulEEG approach \cite{kumar2022mulEEG}, considering the richness of the deep EEG representation that the model provides by combining SSL with a multi-view description. We differentiate from the original model with respect to the focus on computational efficiency and in solving a real-world task, which is classification, instead of feature representation learning, which is the focus in \cite{kumar2022mulEEG}.

	\section{BACKGROUND}
	\label{sec:background}
	In this section, the fundamental elements to be used in our methodology are described. They are the mulEEG, a multi-view representation learning model on EEG signals \cite{kumar2022mulEEG}, and the Linear Support Vector Machine \cite{bishop2006pattern}, which is chosen here as the classification head.

        \subsection{mulEEG}
        
	The mulEEG model, presented in \cite{kumar2022mulEEG}, is a self-supervised multi-view method to learn the representation of EEG signals. The objective is to effectively utilize the complementary information of the EEG multi-view signals. Also, this self-supervised method follows the contrastive learning approach.
 
	In order to obtain the multi-view of EEG signals, first data augmentation is applied. The family of augmentations $T_1$ uses jittering, in which uniform noise is added to the EEG signal, and masking, in which signals are masked randomly, ending up with the time series $t_1$. On the other hand, in the family of augmentations $T_2$, the EEG signals are randomly flipped in horizontal direction and then scaled with Gaussian noise, resulting in the time series $t_2$. Additionally, $t_1$ and $t_2$ are converted into their respective spectrograms, $s_1$ and $s_2$, by a Short-Time Fourier Transform $S$. In this way, we have all the EEG signal views to be used.

	Thereby, the mulEEG model is composed by the time series encoder $E_t$, which is a ResNet-50 with 1D convolutions, and the spectrogram encoder $E_s$ as feature extractors. They are responsible for obtaining the effective EEG signal representations.
 
\begin{figure*}[!htpb]
        \centering
        \includegraphics[width=0.85\linewidth]{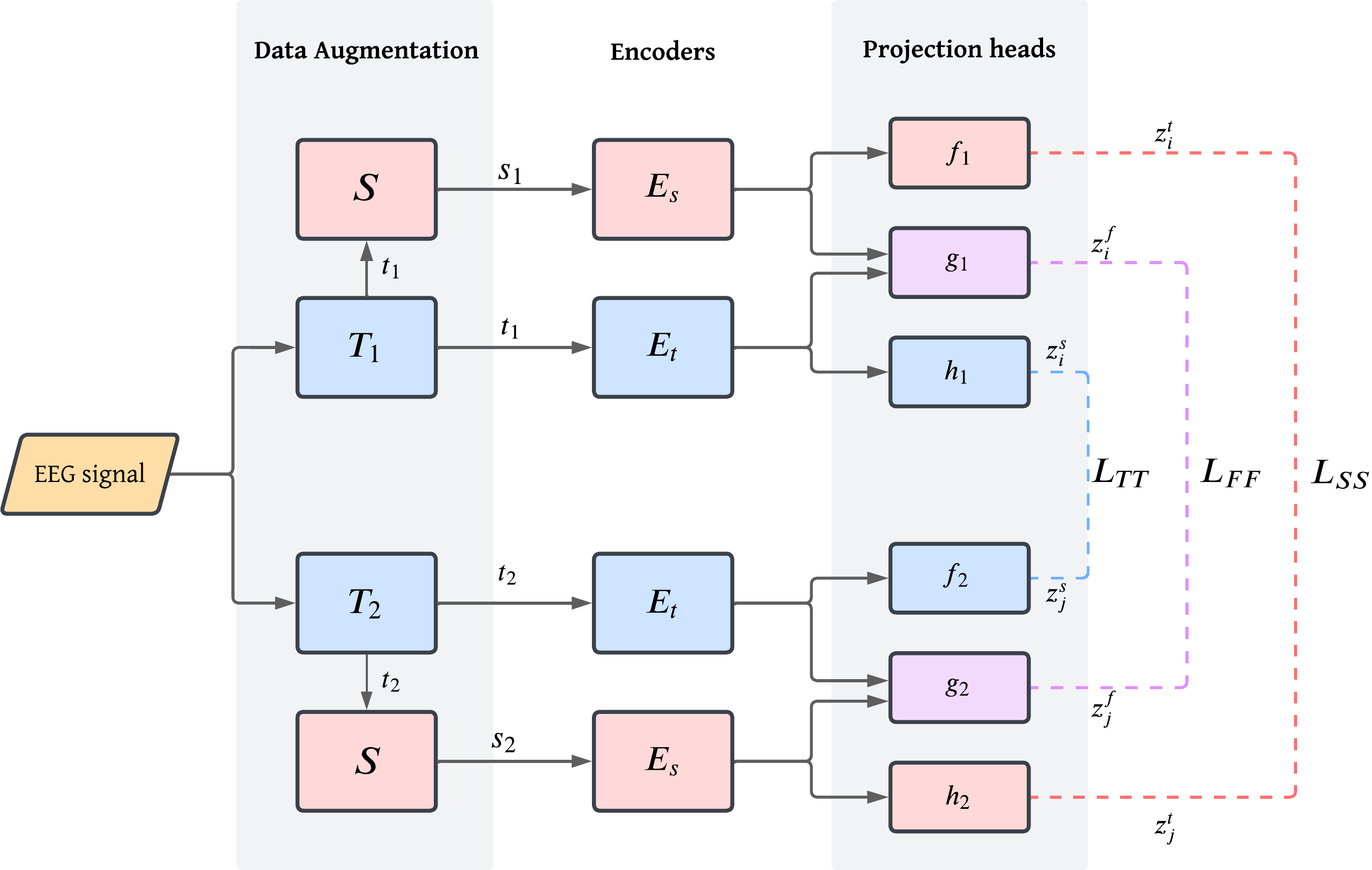}
        \caption{MulEEG structure. First, the EEG signal goes through data augmentation, in which we get the time series views of the families of augmentations $T_1$ and $T_2$ and their conversions to spectrograms by the Short Time Fourier Transform $S$. Then, these views are passed to encoders $E_t$ for time series and $E_s$ for spectrograms. Finally, the outputs go through distinct projection heads depending on the type of view (time series, spectrogram or their concatenation) and the family of augmentations, where the contrastive loss is applied between the projections of the same type.}
        \label{fig:mulEEG}
    \end{figure*}

	Once the time series and spectrogram features have been obtained, they are passed into projection heads, which map those representations to the space where the contrastive loss will be applied. This structure is a fully connected neural network whose layer sequence corresponds to: Linear, Batch Norm, ReLU, and Linear. For each family of augmentations $T_i$, $i=1,2$, there are three projection heads, one for each kind of feature: time series ($f_i$), spectrogram ($h_i)$, and the concatenation of both ($g_i$). The corresponding contrastive losses are named $L_{TT}$, $L_{SS}$, and $L_{FF}$, which gives more flexibility to optimize each feature. The diagram of mulEEG is presented in Figure \ref{fig:mulEEG}.

    The model employs a variant of contrastive loss called NT-Xent, which maximizes the similarity between two augmented views while minimizing its similarity with other samples \cite{kumar2022mulEEG}, given by Equation (\ref{eq:NT-Xent}). Notice that $N$ is the batch size, $\tau$ is the temperature parameter, and cosine similarity is used.
    \begin{equation}
        \label{eq:loss}
        l(i,j)=-\log \left ( \frac{\exp(\cos(\mathbf{z_i}, \mathbf{z_j})/\tau)}{\sum_{k=1}^{2N} \mathbb{I}_{[k \neq i]} \exp(\cos(\mathbf{z_i}, \mathbf{z_k})/\tau)} \right ),
    \end{equation}
    \begin{equation}
        \label{eq:NT-Xent}
        L(\mathbf{z_i},\mathbf{z_j})= \frac{1}{2N} \sum_{k=1}^N l(2k-1,2k)+l(2k,2k-1).
    \end{equation}
    
    We also have the diverse loss $L_D$, which forces the complementary information between time series and spectrogram views. This loss is applied over the time series and spectrogram features from both families of augmentations of a single sample instead of the entire batch, ignoring the concatenated features, which tend to maximize the mutual information between those features. The diverse loss is represented in Equation (\ref{eq:diverse-loss}), where $z_k=[\mathbf{z_i^t,z_j^t,z_i^s,z_j^s}]$ is taken with respect to a single sample and $\tau_d$ is the temperature parameter. The total loss, a linear combination of all the losses previously described with parameters $\lambda_1$ and $\lambda_2$, is represented by Equation \ref{eq:total-loss}.
    \begin{equation}
        \scalebox{0.95}{
        $l_d(z_k,a,b)=-\log \left ( \frac{\exp(\cos(z_k[a], z_k[b])/\tau_d)}{\sum_{i=1}^{4} \mathbb{I}_{[i \neq a]} \exp(\cos(z_k[a],z_k[i])/\tau_d)} \right ),$}
    \end{equation}
    \begin{equation}
    \label{eq:diverse-loss}
        \scalebox{0.8}{
        $L_D = \frac{1}{4N} \sum_{k=1}^N l_d(z_k,1,2)+l_d(z_k,2,1) + l_d(z_k,3,4) + l_d(z_k,4,3),$}
    \end{equation}
    \begin{equation}
    \label{eq:total-loss}
        L_{tot} = \lambda_1 (L_{TT} + L_{FF} + L_{SS}) + \lambda_2 L_D.
    \end{equation}

\begin{figure}[!htpb]
    \centering
    \includegraphics[width=\linewidth]{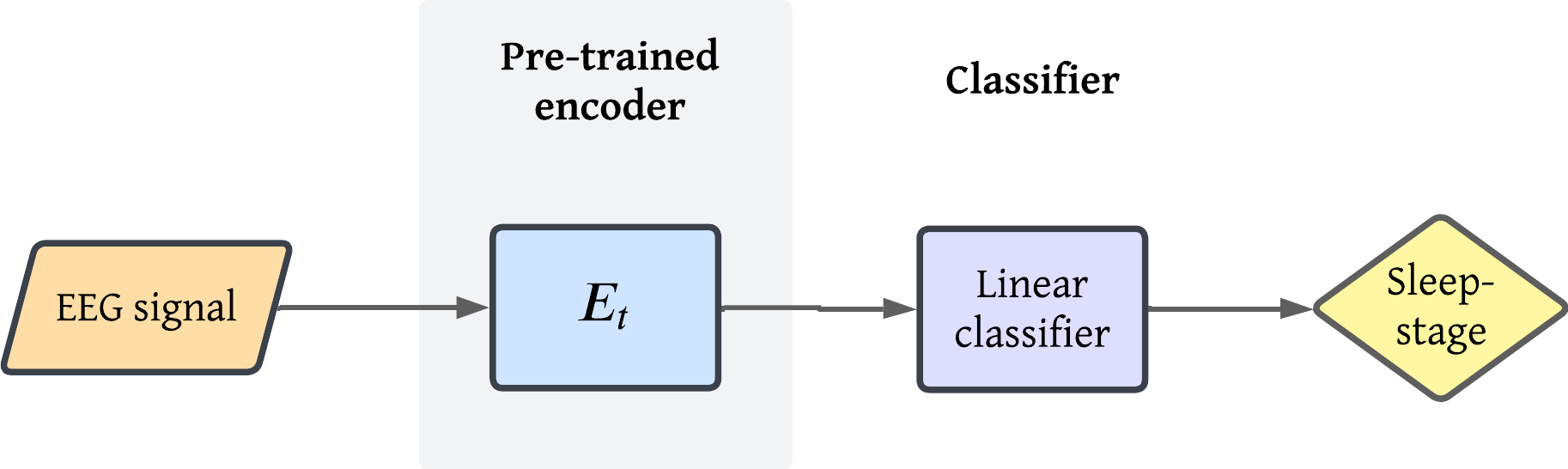}
    \caption{Linear evaluation of mulEEG, in which the EEG signal goes through the pre-trained encoder $E_t$ and the representation obtained is used as input to train the linear classifier, getting the sleep-stage classification at the end.}
    \label{fig:linear-evaluation}
\end{figure}

 	For the final outcome of the model, the pre-trained encoders are submitted to a linear layer, which is attached right after the frozen time series encoder and only this specific layer is trained, evaluating the chosen metrics as shown in Figure \ref{fig:linear-evaluation}. Therefore, mulEEG does not explore the sleep-stage classification task but learns effective representations of it.

	\subsection{Linear SVM}
	Support Vector Machine (SVM) is a supervised machine learning algorithm that can be used in binary classification problems. The SVM goal is to obtain the maximum margin that separates hyperplanes corresponding to decision boundaries over the classes. Next, the Linear SVM description is presented according to \cite{scikit-learn}.
 
 	Given a sample $(\mathbf{x}, y)$, with input vector $\mathbf{x} \in \mathbb{R}^{n \times 1}$ and label $y \in \{-1,1\}$, consider the prediction given by $\mbox{sign}(\mathbf{w}^T \mathbf{x} + b)$, where $\mathbf{w} \in \mathbb{R}^{n \times 1}$ is the weight and $b \in \mathbb{R}$ is the bias term. The hinge loss used in Linear SVM can be defined as
 \begin{equation}
 \label{hinge-loss}
     \max (0, 1-y(\mathbf{w}^T \mathbf{x} + b)).
 \end{equation}
 
 \noindent Note that if the prediction is correct, the hinge loss is equal to zero. 
 
  	Thereby, given $m$ samples $\{(\mathbf{x^{(i)}}, y^{(i)})\}_{i=1}^m$, the Linear SVM solves the following problem:
 \begin{equation}
 \label{cost-SVM}
     \underset{\mathbf{w}, b}{\arg\min} \left ( C \sum_{i=1}^m \max (0, 1-y^{(i)}(\mathbf{w}^T \mathbf{x^{(i)}} + b)) + \frac{1}{2} \mathbf{w}^T \mathbf{w} \right ).
 \end{equation}

 \noindent In Equation (\ref{cost-SVM}), a regularization term is included via $C>0$, which acts as the inverse of the regularization penalty.

  	One can also increase the complexity of SVM by the addition of a kernel function, which transforms the input vectors in such a way that the class separation is performed by complicated non-linear decision boundaries. However, in the Linear SVM, the kernel function is the identity. Additionally, in a multi-class classification problem, strategies like ``one-vs-the-rest'' or ``one-vs-one'' can be applied.

 
	\section{PROPOSED METHOD}
	\label{sec:proposed}
	
	Taking into consideration that, despite its effectiveness in EEG analysis, mulEEG is a complex model and its training algorithm has a high computational cost, the proposed method adapts the original architecture, with the objective of obtaining similar accuracy with a singnificantly reduced computational overhead. In a first stage, we substitute the ResNet-50 in the time series encoder by a ResNet-18, also using 1D-convolutions. This adaptation aims to verify whether a simpler model can also be effective in the classification task. The amount of data used to train the model was also varied to check the need for a large dataset for this task. The architectures for ResNet-50 and ResNet-18 with 1D convolutions in the encoder $E_t$ are presented in Tables \ref{tab: resnet-50_1D} and \ref{tab: resnet-18_1D}, respectively. The architecture of the ResNet-18 was adapted with 1D-convolutions based on \cite{ResNet}.

\begin{table*}[!htpb]
\caption{Architecture of ResNet-50 with 1D-convolutions used in mulEEG. Consider the input dimension as $[256 \times 1 \times 3000]$, where $256$ is the batch size, $1$ is the number of channels, and $3000$ is the length of each sample. The residual blocks are represented in brackets and their numbers of repetition are indicated at right. Additionally, $k$ is the kernel size, $f$ is the number of filters, $s$ is the stride, and $p$ is the padding of the 1D-convolution. Notice that $s=2$ means that the stride is actually equal to $2$ only in the first repetition of the corresponding bottleneck block, so there is a down-sampling. In the next repetitions, the stride is equal to $1$.}

\centering
\begin{tabular}{ccc}
\multicolumn{3}{c}{ResNet-50 with 1D-convolutions used in mulEEG.}                                                                                                                                  \\ \hline
Layer         & Output dimension             & Architecture                                                                                                              \\ \hline
\multirow{2}{*}{Conv0} & $[256 \times 16 \times 1500]$ & $k=71$, $f=16$, $s=2$, $p=35$                                                                                            \\
                       & $[256 \times 16 \times 750]$  & $k=71$, $s=2$, $p=35$ Max-Pooling                                                                                        \\ \hline
Conv1\_x               & $[256 \times 32 \times 750]$  & $\begin{bmatrix}k=1 & f=8 & s=1 & p=0 \\ k=25 & f=8 & s=1 & p=12 \\ k=1 & f=32 & s=1 & p=0 \\ \end{bmatrix} \times 3$    \\ \hline
Conv2\_x               & $[256 \times 64 \times 375]$  & $\begin{bmatrix}k=1 & f=16 & s=1 & p=0 \\ k=25 & f=16 & s=2 & p=12 \\ k=1 & f=64 & s=1 & p=0 \\ \end{bmatrix} \times 4$  \\ \hline
Conv3\_x               & $[256 \times 128 \times 188]$ & $\begin{bmatrix}k=1 & f=32 & s=1 & p=0 \\ k=25 & f=32 & s=2 & p=12 \\ k=1 & f=128 & s=1 & p=0 \\ \end{bmatrix} \times 6$ \\ \hline
Conv4\_x               & $[256 \times 256 \times 94]$ & $\begin{bmatrix}k=1 & f=64 & s=1 & p=0 \\ k=25 & f=64 & s=2 & p=12 \\ k=1 & f=256 & s=1 & p=0 \\ \end{bmatrix} \times 3$
\end{tabular}
\label{tab: resnet-50_1D}
\end{table*}

\begin{table*}[!htpb]
\caption{
Architecture of the ResNet-18 adapted in our method with 1D-convolutions. The notation is identical to that in Table \ref{tab: resnet-50_1D}.}
\centering
\begin{tabular}{ccc}
\multicolumn{3}{c}{ResNet-18 adapted with 1D-convolutions}                                                                                                                                  \\ \hline
Layer         & Output dimension             & Architecture                                                                                                            \\ \hline
\multirow{2}{*}{Conv0} & $[256 \times 16 \times 1500]$ & $k=71$, $f=16$, $s=2$, $p=35$                                                                                            \\
                       & $[256 \times 16 \times 750]$  & $k=71$, $s=2$, $p=35$ Max-Pooling                                                                                        \\ \hline
Conv1\_x               & $[256 \times 8 \times 750]$   & $\begin{bmatrix}k=25 & f=8 & s=1 & p=12 \\ k=25 & f=8 & s=1 & p=12 \\ \end{bmatrix} \times 2$    \\ \hline
Conv2\_x               & $[256 \times 16 \times 375]$  & $\begin{bmatrix}k=25 & f=16 & s=1 & p=12 \\ k=25 & f=16 & s=2 & p=12 \\ \end{bmatrix} \times 2$    \\ \hline
Conv3\_x               & $[256 \times 32 \times 188]$  & $\begin{bmatrix}k=25 & f=32 & s=1 & p=12 \\ k=25 & f=32 & s=2 & p=12 \\ \end{bmatrix} \times 2$    \\ \hline
Conv4\_x               & $[256 \times 64 \times 94]$   & $\begin{bmatrix}k=25 & f=64 & s=1 & p=12 \\ k=25 & f=64 & s=2 & p=12 \\ \end{bmatrix} \times 2$    \\ \hline
\end{tabular}
\label{tab: resnet-18_1D}
\end{table*}

\begin{figure}[!htpb]
    \centering
    \includegraphics[width=\linewidth]{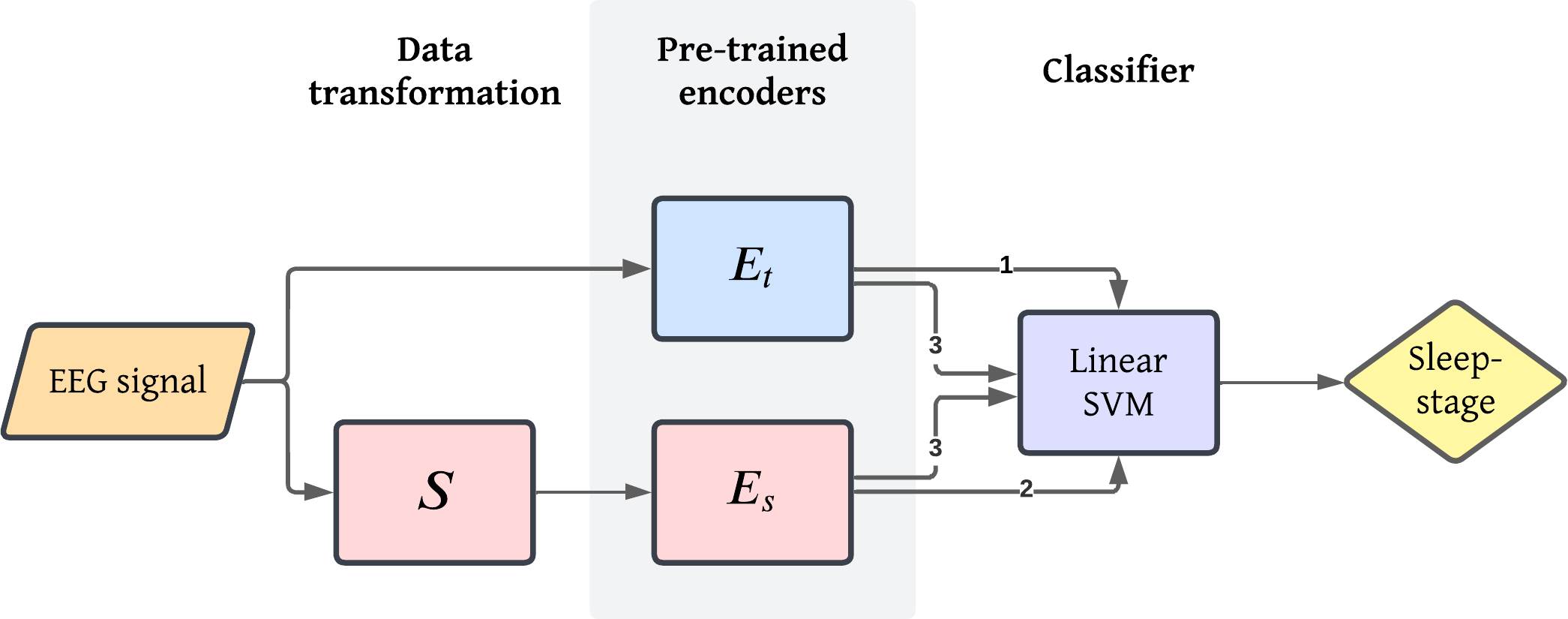}
    \caption{Proposed method. First, the EEG signal goes through the chosen pre-trained encoders, which could vary depending on the model for $E_t$. Notice that before $E_s$, there is the Short Time Fourier Transform $S$. Then, the Linear SVM is trained with one of the possible inputs: (1) time series representation, (2) spectrogram representation or (3) their concatenation. Finally, the sleep-stage classification is obtained.}
    \label{fig:proposed-method}
\end{figure}

	Additionally, once the EEG signal representations were learned, a Linear SVM is trained for sleep-stage classification. This step has the goal of checking how effective those features are in a more robust classification algorithm, since the baseline architecture had a simple linear classifier to provide its output. The choice of using a linear kernel on SVM is justified by the huge data volume used in the experiments. According to \cite{scikit-learn}, an SVM with non-linear kernel scales at least quadratically with the number of samples, while the SVM with linear kernel can scale almost linearly to millions of samples. Thereby, the Linear SVM is a sufficient algorithm for the goal of testing the learned EEG features in the classification task. Finally, in the proposed framework, the input of the Linear SVM can be the time series features, spectogram features, or the concatenation of both, as shown in Figure \ref{fig:proposed-method}.
    
	\section{EXPERIMENTAL SETUP}
	\label{sec:experiments}

	\subsection{Dataset}
	The proposed method was evaluated on the Sleep-EDF database presented in \cite{Kemp-etal-2000} and publicly available in \cite{Goldberger-etal-2000}. The data consists of $153$ whole-night polysomnography sleep recordings sampled at $100$ Hz and by the EEG method (from Fpz-Cz and Pz-Oz electrode locations), presented in the *PSG.edf files. Each one contains the respective *Hypnogram.edf file with annotations of the sleep patterns (hypnograms), which are the sleep stages `W' (Wake), `R' (REM), `1' (N1), `2' (N2), `3' (N3), `4' (N3), `M' (moviment time) and `?' (not scored) scored by well-trained technicians. These data come from the Sleep Cassette Study conducted between $1987$ and $1991$, which is about the age effects on sleep in healthy Caucasian adults aged 25-101, without any sleep-related medication \cite{Goldberger-etal-2000}. 

	In terms of data processing, as in \cite{kumar2022mulEEG}, $58$ patients were chosen to compose the unlabeled pretext group for training the mulEEG and $20$ were left for cross-validation (5-fold) in the linear evaluation. Each recording, sampled at $100$ Hz, was split into $30$-second segments named epoch, which corresponds to a $3000$ components array. To train the model, we randomly select $20\%$ and $100\%$ of those samples from the pretext group. Also, the same data used in the linear evaluation was used to train the Linear SVM, except that in this last case they are normalized.
	
	\subsection{Implementation Details}
	 
	In general, the training protocol was kept similar to that in \cite{kumar2022mulEEG}. Then, we have $N=256$ (batch size), the temperature parameters $\tau=1$ for $L_{TT}$, $L_{FF}$, and $L_{SS}$ and $\tau_d=10$ for $L_d$, $\lambda_1=1$, and $\lambda_2=2$. An important difference in the protocol was with respect to the number of epochs, which here was defined as $200$, while the authors in \cite{kumar2022mulEEG} used $140$. From epoch $80$ on, the linear evaluation starts to be done at each $4$ epochs, in which the linear classifier is trained for $100$ epochs with the corresponding pre-trained encoder for time series $E_t$ frozen. The computational setup for training comprised an Intel Core-i7 8700, 16 GB of RAM, Nvidia Titan V graphics card, Python 3.9/PyTorch 2.5, Linux Ubuntu 24.10.

    In the Linear SVM, the loss function is the squared hinge loss, with $C=1$, tolerance for stopping equals $1^{-4}$, and the maximum number of iterations equals $1000$. The pre-trained encoders used for this task are the ones with the highest MF1 during trainig with $100\%$ of data and varyng the ResNet architectures of $E_t$.

	\subsection{Evaluation Metrics}
	
    The evaluation metrics used for both the linear evaluation of mulEEG and the proposed method are accuracy (Acc), Cohen's kappa ($\kappa$), and macro-averaged F1 score (MF1). The computational time for training was also analyzed.
	
	\section{RESULTS AND DISCUSSION}
	\label{sec:results}

Table \ref{tab:mulEEG-training} presents the linear evaluation metrics for various mulEEG training configurations compared to \cite{kumar2022mulEEG}. It should be noted that the use of ResNet-50 results in significantly longer training time compared to ResNet-18, despite delivering superior results. However, the metrics obtained with ResNet-50 and $100\%$ of the pretext group were the best in all experiments, although slightly lower than those reported in \cite{kumar2022mulEEG}. However, this configuration proved to be the most computationally expensive. Conversely, training with ResNet-18 and $100\%$ of the data produced metrics very similar to those achieved with ResNet-50 and $20\%$ of the data, with the latter one requiring approximately one-third of the training time. In this way, we confirm that training with ResNet-50 and $20\%$ of the data provides the best cost-benefit ratio. In other words, reducing the data volume of the pretext group is significantly more advantageous than simplifying the model to accelerate training.
    
\begin{table*}[!htpb]
\caption{Evaluation metrics of various mulEEG training configurations.}
\centering
\begin{tabular}{|l|c|c|c|c|c|}
\hline
\multicolumn{1}{|c|}{\textbf{Method}} & \textbf{Acc} & \textbf{$\kappa$} & \textbf{MF1}       & \textbf{Training time}\\ \hline
20\% data + ResNet-18            & 0.6979            & 0.5705                  & 0.5252  & 3h 2m 55s     \\  
20\% data + ResNet-50            & 0.7483            & 0.6469                  & 0.6056  & 4h 27m 26s     \\  
100\% data + ResNet-18           & 0.7549            & 0.6528                  & 0.6189  & 13h 21m 22s      \\
100\% data + ResNet-50           & 0.7653          & 0.6704                  & 0.6546    & 21h 2m 56s                                                      \\ \hline
Linear evaluation of \cite{kumar2022mulEEG}         & 0.7806   & 0.6850         & 0.6782  & -                                      \\ \hline
\end{tabular}
\label{tab:mulEEG-training}
\end{table*}	

For the linear SVM classification task, the metrics obtained by varying the time series encoder and classifier input are presented in Table \ref{tab:SVM-training}. First, it is evident that the metrics obtained by SVM when using the EEG signal were significantly inferior compared to the other configurations. This suggests that the representations learned by mulEEG are indeed effective for the classification task, serving as a proficient feature extractor for EEG signals.

\begin{table*}[!htpb]
\caption{Metrics observed in the classification task using Linear SVM with EEG signal and time series, spectrogram, and concatenated features as the input. The encoder $E_t$ was varied and the linear evaluation of \cite{kumar2022mulEEG} is also presented for comparison purposes.}
\centering
\begin{tabular}{|clccc|}
\hline
\multicolumn{5}{|c|}{\textbf{Linear SVM training}}                                                                                                                                                        \\
\hline
\multicolumn{1}{|l|}{\textbf{$E_t$ model}} & \multicolumn{1}{l|}{\textbf{Input of SVM}} & \multicolumn{1}{c|}{\textbf{Acc}}    & \multicolumn{1}{c|}{\textbf{$\kappa$}} & \textbf{MF1}           \\ \hline
\multicolumn{2}{|l|}{Linear evaluation of \cite{kumar2022mulEEG}}                & \multicolumn{1}{c|}{0.7806} & \multicolumn{1}{c|}{0.6850}   & 0.6782  \\
\hline
\multicolumn{1}{|c|}{-}                           & \multicolumn{1}{l|}{EEG signal}         & \multicolumn{1}{c|}{0.2984}          & \multicolumn{1}{c|}{0.0260}            & 0.2143                \\ \hline
\multicolumn{1}{|c|}{\multirow{3}{*}{ResNet-18}}  & \multicolumn{1}{l|}{Time series feature}        & \multicolumn{1}{c|}{0.7732}          & \multicolumn{1}{c|}{0.6812}            & 0.6657                \\
\multicolumn{1}{|c|}{}                            & \multicolumn{1}{l|}{Spectrogram feature} & \multicolumn{1}{c|}{0.7452}          & \multicolumn{1}{c|}{0.6415}            & 0.6426                \\
\multicolumn{1}{|c|}{}                            & \multicolumn{1}{l|}{Concatenated feature}     & \multicolumn{1}{c|}{0.7909}          & \multicolumn{1}{c|}{0.7074}            & 0.6972                \\ \hline
\multicolumn{1}{|c|}{\multirow{3}{*}{ResNet-50}}  & \multicolumn{1}{l|}{Time series feature}        & \multicolumn{1}{c|}{0.8090}          & \multicolumn{1}{c|}{0.7328}            & 0.7239                \\
\multicolumn{1}{|c|}{}                            & \multicolumn{1}{l|}{Spectrogram feature} & \multicolumn{1}{c|}{0.7413}          & \multicolumn{1}{c|}{0.6357}            & 0.6366                \\
\multicolumn{1}{|c|}{}                            & \multicolumn{1}{l|}{Concatenated feature}     & \multicolumn{1}{c|}{0.8079}          & \multicolumn{1}{c|}{0.7323}            & 0.7373                \\ \hline
\end{tabular}
\label{tab:SVM-training}
\end{table*}

It is also noteworthy that when using spectrogram features, the metrics remain similar despite variations in the encoder. This aligns with expectations, as the spectrogram encoder $E_s$ remains unchanged in both cases, indicating that its training is unaffected by the time series encoder, as intended by the use of $L_{SS}$ loss function. However, the use of this feature proved to be the least beneficial to the classification task, outperforming only the direct use of the EEG signal.

Furthermore, when using ResNet-18, the time series feature yields metrics that are highly comparable with those reported in \cite{kumar2022mulEEG}, while the concatenated features surpass the metrics of the aforementioned reference. These results demonstrate that simplifying the mulEEG model by using ResNet-18 in $E_t$ is highly advantageous for classifying EEG signals with a more robust classifier, particularly when utilizing concatenated features, demonstrating the effective use of the complementary information from both views.

Moreover, when employing ResNet-50, i.e., mulEEG in its original configuration, the use of both time series and concatenated features yields nearly identical results, surpassing all other metrics, including those of \cite{kumar2022mulEEG}. Unlike the findings with ResNet-18, there is no significant improvement when using the concatenated features, with only the F1 score showing an increase of approximately $2\%$. From this we can infer that as the complexity of the temporal encoder increases, more information is extracted from the time series, making the spectrogram feature less contributory to the classifier's learning process.

In conclusion, the feature extractors trained within mulEEG play a pivotal role in the classification of EEG signals into sleep stages. Furthermore, it can be posited that the superior performance of the SVM in classifying EEG signals arises from using the concatenated features, which outperform the linear evaluation presented in \cite{kumar2022mulEEG}. Despite the superior metrics of using ResNet-50 as the encoder in this context, ResNet-18 still offers a compelling cost-benefit ratio when paired with a more robust classifier. After all, a more streamlined model achieved competitive performance by effectively exploiting the complementary information between the time series and spectrogram. {\color{black}This also confirms that both data representations offer useful viewpoints and that the combination of a self-supervised feature description allows the effective use of lighter supervised models in the target task. This finding can also help in other domains to be explored in future works, for time series classification in general.}


	
	\section{CONCLUSIONS}
	\label{sec:conclusions}

This work presents an investigation on the use of a self-supervised model alongside with Linear SVM classifier, with the aim of exploring the classification of EEG signals into sleep-stages. Initially, we observed that the complexity of deep learning models known to be well-succeeded in this task, combined with the large volume of data, resulted in a high computational cost during training. To address this, we propose a model that uses as baseline the well-established mulEEG architecture, but simplifies it by replacing the ResNet-50 with 1D-convolutions by ResNet-18, using it as the time series encoder. Furthermore, training this model involved experimenting with both $20\%$ and $100\%$ of the data from the pretext group. The linear evaluation of these training configurations revealed that although the ResNet-50 model with $100\%$ of the data achieved the best metrics, it also incurred in the highest computational cost. We thus discovered that reducing the volume of data yielded a better cost-benefit ratio than simplifying the temporal encoder model.

It is important to notice that the primary goal of models like mulEEG is to learn effective representations of EEG signals, rather than to directly address the classification task. This is evident in the linear evaluation, where a simple linear classifier receives only the time series features as input. In this context, the Linear SVM was chosen here as a more robust classifier, trained with varying inputs derived from the EEG signal representations obtained by the self-supervised model, which receives the time series, spectrogram, and their concatenation as input. Additionally, the temporal encoder was again varied between ResNet-50 and ResNet-18 in order to compare their performance with a more robust classifier. The metrics obtained with ResNet-50, using both time series and concatenated features, were very similar and resulted in the best performance, surpassing the linear evaluation in \cite{kumar2022mulEEG}. It is worth noting that the complexity of the temporal encoder limits the amount of complementary information provided by the spectrogram features. In contrast, when using ResNet-18, the concatenated features emerge as particularly effective, and the results once again outperform those in \cite{kumar2022mulEEG}. This highlights the significant cost-benefit advantage of simplifying the mulEEG model and pairing it with a more robust classifier, thereby effectively capitalizing on the complementary perspectives of both the raw time series and the spectrogram.

In conclusion, this study highlights the potential of associating a robust classifier after extracting features from EEG signals for classification into sleep stages, as well as leveraging the different perspectives these data provide. Indeed, this approach holds promise for further exploration in a variety of problems involving biological signals in general, such as the detection of anomalies in electrocardiograms, for example.

\section*{\uppercase{Acknowledgements}}

J. B. Florindo gratefully acknowledges the financial support of the S\~ao Paulo Research Foundation (FAPESP) (Grants \#2024/01245-1 and \#2020/09838-0) and from National Council for Scientific and Technological Development, Brazil (CNPq) (Grant \#306981/2022-0).


\begin{thebibliography}{}
	
	\bibitem[Bishop and Nasrabadi, 2006]{bishop2006pattern}
	Bishop, C.~M. and Nasrabadi, N.~M. (2006).
	\newblock {\em Pattern recognition and machine learning}, volume~4.
	\newblock Springer.
	
	\bibitem[Eldele et~al., 2023]{eldele2023self}
	Eldele, E., Ragab, M., Chen, Z., Wu, M., Kwoh, C.-K., and Li, X. (2023).
	\newblock Self-supervised learning for label-efficient sleep stage
	classification: A comprehensive evaluation.
	\newblock {\em IEEE Transactions on Neural Systems and Rehabilitation
		Engineering}, 31:1333--1342.
	
	\bibitem[Goldberger et~al., 2000]{Goldberger-etal-2000}
	Goldberger, A., Amaral, L., Glass, L., Hausdorff, J., Ivanov, P.~C., Mark, R.,
	Mietus, J.~E., Moody, G.~B., Peng, C.~K., and Stanley, H.~E. (2000).
	\newblock Physiobank, physiotoolkit, and physionet: Components of a new
	research resource for complex physiologic signals.
	\newblock {\em Circulation [Online]}, 101 (23):pp. e215--e220.
	
	\bibitem[He et~al., 2015]{ResNet}
	He, K., Zhang, X., Ren, S., and Sun, J. (2015).
	\newblock Deep residual learning for image recognition.
	\newblock {\em CoRR}, abs/1512.03385.
	
	\bibitem[Kemp et~al., 2000]{Kemp-etal-2000}
	Kemp, B., Zwinderman, A.~H., Tuk, B., Kamphuisen, H. A.~C., and Oberye, J.
	J.~L. (2000).
	\newblock Analysis of a sleep-dependent neuronal feedback loop: the slow-wave
	microcontinuity of the eeg.
	\newblock {\em IEEE Transactions on Biomedical Engineering}, 47(9):1185--1194.
	
	\bibitem[Kumar et~al., 2022]{kumar2022mulEEG}
	Kumar, V., Reddy, L., Sharma, S.~K., Dadi, K., Yarra, C., Bapi, R.~S., and
	Rajendran, S. (2022).
	\newblock muleeg: A multi-view representation learning on eeg signals.
	\newblock {\em Lecture Notes in Computer Science}, 13433.
	
	\bibitem[Liu et~al., 2024]{liu2024automatic}
	Liu, P., Qian, W., Zhang, H., Zhu, Y., Hong, Q., Li, Q., and Yao, Y. (2024).
	\newblock Automatic sleep stage classification using deep learning: signals,
	data representation, and neural networks.
	\newblock {\em Artificial Intelligence Review}, 57(11):301.
	
	\bibitem[Patel et~al., 2024]{sleep-stages}
	Patel, A.~K., Reddy, V., Shumway, K.~R., and Araujo, J.~F. (2024).
	\newblock Physiology, sleep stages.
	
	\bibitem[Pedregosa et~al., 2011]{scikit-learn}
	Pedregosa, F., Varoquaux, G., Gramfort, A., Michel, V., Thirion, B., Grisel,
	O., Blondel, M., Prettenhofer, P., Weiss, R., Dubourg, V., Vanderplas, J.,
	Passos, A., Cournapeau, D., Brucher, M., Perrot, M., and Duchesnay, E.
	(2011).
	\newblock Scikit-learn: Machine learning in {P}ython.
	\newblock {\em Journal of Machine Learning Research}, 12:2825--2830.
	
	\bibitem[Sekkal et~al., 2022]{SEKKAL2022103751}
	Sekkal, R.~N., Bereksi-Reguig, F., Ruiz-Fernandez, D., Dib, N., and Sekkal, S.
	(2022).
	\newblock Automatic sleep stage classification: From classical machine learning
	methods to deep learning.
	\newblock {\em Biomedical Signal Processing and Control}, 77:103751.
	
	\bibitem[Xiao et~al., 2024]{xiao2024self}
	Xiao, T., Wang, Z., Zhang, Y., Wang, S., Feng, H., Zhao, Y., et~al. (2024).
	\newblock Self-supervised learning with attention mechanism for eeg-based
	seizure detection.
	\newblock {\em Biomedical Signal Processing and Control}, 87:105464.
	
	\bibitem[Yuan et~al., 2024]{yuan2024self}
	Yuan, H., Plekhanova, T., Walmsley, R., Reynolds, A.~C., Maddison, K.~J.,
	Bucan, M., Gehrman, P., Rowlands, A., Ray, D.~W., Bennett, D., et~al. (2024).
	\newblock Self-supervised learning of accelerometer data provides new insights
	for sleep and its association with mortality.
	\newblock {\em NPJ digital medicine}, 7(1):86.
	
\end{thebibliography}

\end{document}